%% file: main.tex
\documentclass{article}

\usepackage{microtype}
\usepackage{graphicx}
\usepackage{subfigure}
\usepackage{booktabs} 

\usepackage{hyperref}

\usepackage[accepted]{icml2024}

\usepackage{amsmath}
\usepackage{amssymb}
\usepackage{mathtools}
\usepackage{amsthm}
\usepackage{graphicx}
\usepackage{multirow}
\usepackage{xspace}
\usepackage{enumitem}

\usepackage[capitalize,noabbrev]{cleveref}

\theoremstyle{plain}

\theoremstyle{definition}

\theoremstyle{remark}

\usepackage[textsize=tiny]{todonotes}

\newcommand{\ours}{\textsc{MemoryLLM}\xspace}
\icmltitlerunning{\ours: Towards Self-Updatable Large Language Models}

\newcommand{\wy}[1]{{\color{black}{#1}}}

\begin{document}

\twocolumn[
\icmltitle{\ours: Towards Self-Updatable Large Language Models}




\renewcommand{\thefootnote}{\textasteriskcentered}

\begin{icmlauthorlist}
\icmlauthor{Yu Wang\footnote{Work done during the internship at Amazon.}}{ucsd}
\icmlauthor{Yifan Gao}{amazon}
\icmlauthor{Xiusi Chen}{ucla}
\icmlauthor{Haoming Jiang}{amazon}
\icmlauthor{Shiyang Li}{amazon}
\icmlauthor{Jingfeng Yang}{amazon}
\icmlauthor{Qingyu Yin}{amazon}
\icmlauthor{Zheng Li}{amazon}
\icmlauthor{Xian Li}{amazon}
\icmlauthor{Bing Yin}{amazon}
\icmlauthor{Jingbo Shang}{ucsd}
\icmlauthor{Julian McAuley}{ucsd}
\end{icmlauthorlist}

\icmlaffiliation{ucsd}{UC, San Diego}
\icmlaffiliation{ucla}{UC, Los Angeles}
\icmlaffiliation{amazon}{Amazon}

\icmlcorrespondingauthor{Yu Wang}{yuw164@ucsd.edu}
\icmlcorrespondingauthor{Yifan Gao}{yifangao@amazon.com}

\icmlkeywords{memory, large language model}

\vskip 0.3in
]



\printAffiliationsAndNotice{\mbox{}$^*$Work done during the internship at Amazon.}  

\begin{abstract}
Existing Large Language Models (LLMs) usually remain static after deployment, which might make it hard to inject new knowledge into the model.
We aim to 
build models containing a
considerable portion of self-updatable parameters, 
enabling the model to integrate new knowledge effectively and efficiently. 
To this end,
we introduce \ours, a model that comprises a transformer and a fixed-size memory pool within the latent space of the transformer. 
\ours can self-update with text knowledge and memorize the knowledge injected earlier.
Our evaluations demonstrate the ability of \ours to effectively incorporate new knowledge, as evidenced by its performance on model editing benchmarks. 
Meanwhile, the model exhibits long-term information retention capacity, which is validated through our custom-designed evaluations and long-context benchmarks. 
\ours also
shows operational integrity without any sign of performance degradation even after nearly a million memory updates. Our code and model are open-sourced at \url{https://github.com/wangyu-ustc/MemoryLLM}.
\end{abstract}

\input{1_introduction}
\input{3_preliminary}

\input{4_method}
\input{5_experiments}
\input{2_related_work}

\input{6_conclusion}

\clearpage

\input{8_statement}

\bibliography{ref}
\bibliographystyle{icml2024}


\appendix
\onecolumn
\input{7_appendix}

\end{document}

%% file: 1_introduction.tex
\section{Introduction}
\label{sec:introduction}
Despite the impressive performance LLMs demonstrate, a pivotal issue persists: \emph{How should we update the model with the latest knowledge?} Previous solutions can be broadly categorized into three classes: 
\textbf{(1) Retrieval-Based Methods:} These methods rely on information retrieval in a knowledge base~\citep{kNNLM,MemoryBank}. They can yield strong results, but face challenges when redundancy in the knowledge base presents and suffer the logistical issue of managing an ever-expanding repository of knowledge. 
In multi-modality scenarios, retrieval-based methods might require enormous storage space to store all image data (24 images per second for humans) for retrieval purposes. 
\textbf{(2) Model Editing:} This class of methods involves making targeted edits to the model to adapt to new facts while preserving other desired capabilities~\citep{LLMEditing}. Existing methods primarily focus on fact-based editing, which is typically limited to single sentences. 
This limitation becomes more severe when one attempts to inject new knowledge in the form of longer and more complicated contexts. 
\textbf{(3) Long Context Methods:} Another alternative solution is to incorporate all the knowledge into the model's context,  
which essentially makes the context into a knowledge base. 
This differs from retrieval-based methods in that the context directly informs the inference of the model. Methods in this category involve reducing the complexity of attention operations~\citep{SparserTransformer, longformer, linformer}, and modifying positional embeddings~\citep{alibi, lextransformer} to handle longer contexts. However, as complex reasoning tasks are thirsty for massive up-to-date knowledge, the inevitable context overload with long context methods becomes infeasible, as long as the context length is finite.

In response to the challenges identified above, we introduce \ours, a model that embeds a substantial, fixed-size memory pool within its latent space, which serves as the self-updatable parameters.
Specifically, we build the memory pool as hidden vectors within each 
layer of the transformer. 
At each layer, the memory pool contains \textit{memory tokens} representing compressed knowledge.
This design results in a memory pool that is less redundant than traditional knowledge bases in retrieval-based methods or contexts in long-context methods.
To update the memory pool, we devise a \textit{self-update} mechanism to propagate the new knowledge to every layer of the memory. During self-update, 
\ours only updates a proportion of memory in each layer to absorb the incoming knowledge. This allows previously stored knowledge to slowly phase out. 
These designs ensure \ours remains up-to-date while the old knowledge is slowly forgotten. 
After curated training, we update \ours nearly a million times without observing any performance deterioration.

The evaluation of \ours focuses on several key aspects: (1) \textbf{Integration of New Knowledge}: The model's performance is assessed with model editing benchmarks and QA tasks (long context QA benchmarks), where \ours demonstrates substantial improvements over existing methods. 
(2) \textbf{Knowledge Retention Ability}:
\ours is evaluated on long context benchmarks and our knowledge retention experiments, showcasing its ability to recall knowledge. 
(3) \textbf{Robustness}: To test the integrity of the model, we subject \ours to almost a million update steps. 
The results show that our model is functioning properly even after extreme updates. 

In summary, our contributions are as follows:
\begin{itemize}[nosep,leftmargin=*]
    \item We introduce \ours, which features an integrated memory pool within the latent space of an LLM. This memory pool is designed to manage new knowledge integration and encourage minimal information forgetting while being fixed-sized to circumvent the issue of uncontrolled growth.
    \item We augment a 7B parameter model with an extensive memory pool comprising 1B parameters. 
    \item MemoryLLM demonstrates strong performance across various benchmarks, including model editing, long-context evaluation, and our knowledge retention experiments, showcasing its versatility and effectiveness in diverse applications.
\end{itemize}

%% file: 3_preliminary.tex
\section{Preliminaries}
\subsection{Problem Statement}
The primary challenge addressed in this paper is: \emph{How should we design a large language model that is capable of efficiently integrating new knowledge while minimizing the degradation of previously learned knowledge?} 
To make the challenge more specific, we outline several essential properties that we hope to integrate into the new model: \textbf{(1) Effiency}: The process of knowledge injection into the model should be streamlined, potentially eliminating the need for back-propagation for efficiency. \textbf{(2) Efficacy}: It is crucial to ensure that the knowledge is effectively injected into the model, guaranteeing its impact on the model's performance. 
\textbf{(3) Knowledge Retention}: Our model has a fixed-sized memory pool, implying a constant memorization capacity. This necessitates a mechanism for gradually phasing out older knowledge.
\textbf{(4) Integrity}: The model must maintain full functionality regardless of the number of updates made to the memory pool. 
\textbf{(5) Non-redundancy}: We aim for more compact storage of knowledge, reducing redundancy, and optimizing memory usage. 

\subsection{Sketch of \ours}
To address the above challenges, our rough idea is to design a model denoted as $\mathcal{M}_{\theta, \phi}$ consisting of two sets of parameters: $\phi$ and $\theta$. Once we obtain the model, the $\phi$ parameters should be static, while $\theta$ dynamically evolves when encountering new knowledge. This aligns with the intuition that some knowledge within an LLM should never change (persistent truths, encoded by $\phi$) and some knowledge is being updated continuously (fresh information, modeled by $\theta$). 
Specifically, we use an existing large language model (Llama2) to model $\phi$, while $\theta$ is modeled by the memory pool with the detailed structure in Section \ref{ssub:memory_pool}. 
Here we need to design the \textit{self-updating} mechanism of $\theta$ that is pivotal to this process. 
Denoting the new knowledge as $x$, a text paragraph, the \textit{self-updating} process 
refers to 
updating $\theta$ in a way that does not compromise the general capabilities of the model while injecting the latest knowledge $x$ into the memory pool $\theta$ to obtain a new memory pool $\theta'$:
\begin{equation}\label{eq:one_step_injection}
    \theta' = U(\theta, x)
\end{equation}
Here $U$ is the update function which takes the memory pool $\theta$ and the new knowledge $x$ as input and outputs the new memory pool $\theta'$. 
Extending this process to multistep updating, consider a scenario with a never-ending context or a series of conversation histories, represented as $(x_1, \cdots, x_n)$, where $x_i, i\in\{1,\cdots,n\}$ is a text paragraph. The model requires the integration of all these contexts, which can be accomplished using the update function $I$ defined in Eq.(\ref{eq:one_step_injection}): 
\begin{equation}\label{eq:multi_step_injection}
    \theta_n = U(\cdots(U(\theta, x_1), x_n).
\end{equation}
We define the process \textit{self-updating} as modifying the parameters $\theta$ with newly encountered knowledge $x$, essentially enabling the model to read and assimilate knowledge. This design presents two primary challenges: (1) \textbf{Parameter and Interaction Design}: We need to determine the structure for $\theta$ and how it should interact with $\phi$, The goal is to allow the LLM to effectively use the knowledge from the $\theta$ in the generation process. (2) \textbf{Update function design}: It is crucial to design the update function $U$ such that $\theta$ can be updated without disturbing the old knowledge and undermining the overall capabilities of the model.

%% file: 4_method.tex
\section{\ours}
\subsection{Structure Design}
\begin{figure}[t]
\centering
\subfigure[Generation]{\label{fig:generation}\includegraphics[width=0.8\linewidth]{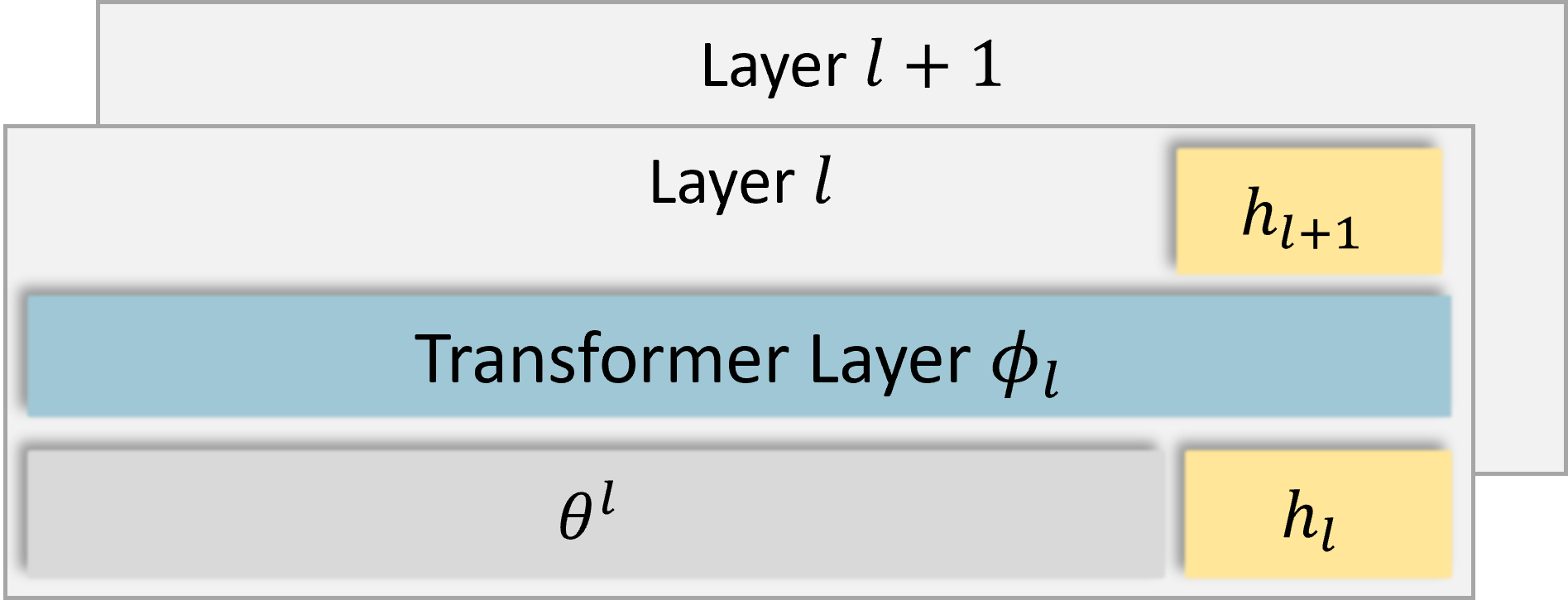}}
\subfigure[Self-Update]{\label{fig:injection}\includegraphics[width=.9\linewidth]{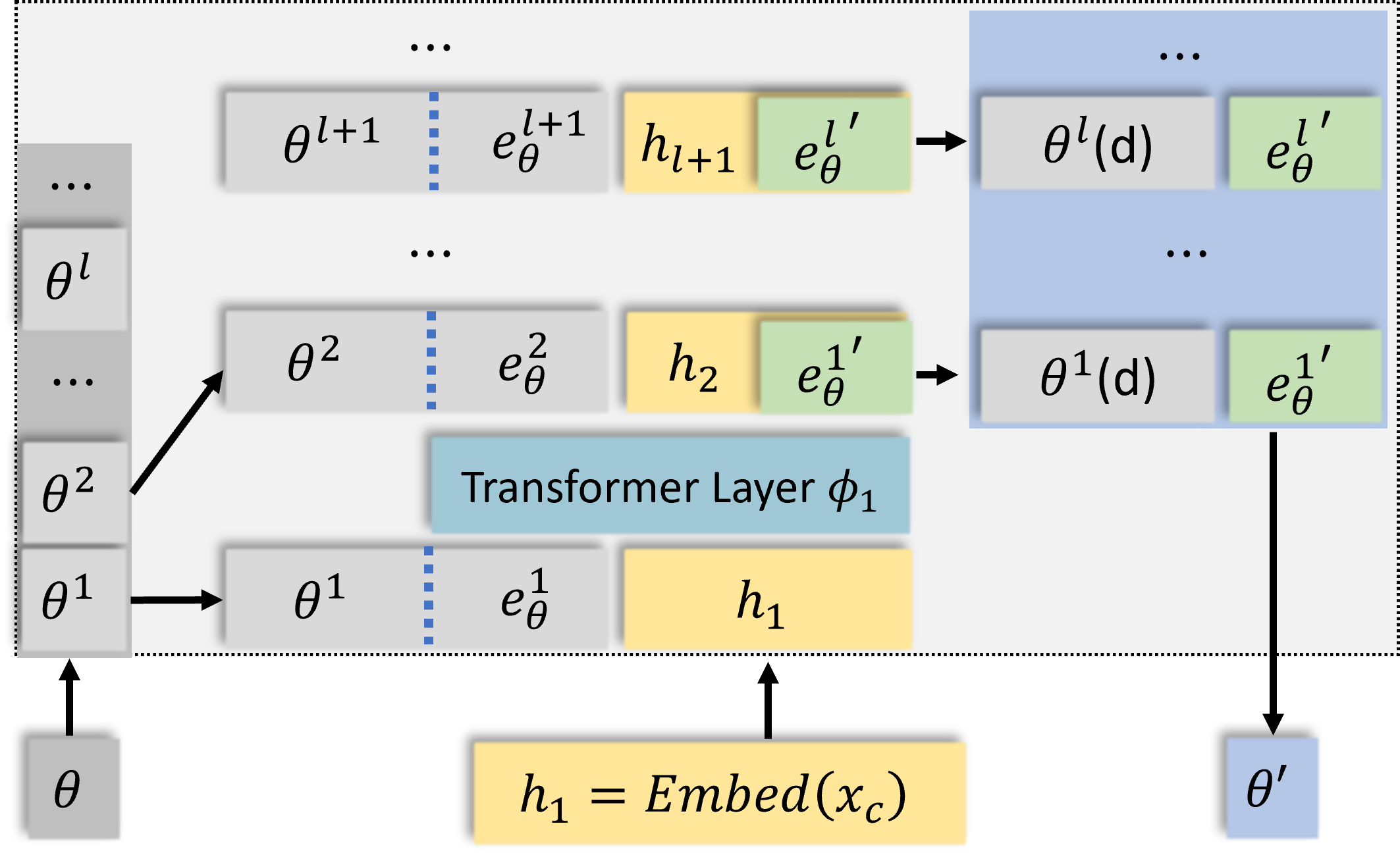}}
\caption{\textbf{The framework of \ours.}
(a) During generation, all memory tokens in the $l$-th layer of memory pool $\theta^l$ are attended by the hidden states $h_l$. (b) During self-update,
The last $k$ memory tokens from $\theta_l$ are taken to be concatenated with the hidden states $h_l$ as the input to $\phi_l$. 
The output $h_{l+1}$ goes to the next layer. The last $K$ tokens of $h_{l+1}$ serve as the new memory tokens ${e_\theta^l}'$. 
We randomly drop $K$ tokens in $\theta_l$ and concatenate the left $\theta_l$ (denoted as $\theta^l(d)$) with ${e_\theta^l}'$ to obtain new memory $\theta_l'$.}
\vspace{-15pt}
\label{fig:model_framework}
\end{figure}

\subsubsection{Memory Pool}
\label{ssub:memory_pool}
We choose to instantiate $\phi$ with an off-the-shelf LLM, specifically Llama2~\citep{llama2}. 
$\phi$ consists of multiple transformer layers, 
denoted as $\phi = \{\phi_l\}_{l=1}^{L}$, where $L$ represents the total number of layers. 
To facilitate the transformer $\phi$ to understand the memory pool $\theta$, we conceptualize $\theta$ as hidden vectors within each transformer layer, symbolized as $\theta = \{\theta_l\}_{l=1}^L$. 
Each $\theta_l$ is of dimension $N \times d$, corresponding to $N$ hidden states and the word embedding dimension $d$ in $\phi$. 
We term $\theta_l$ \textit{memory tokens}. 
The \textit{memory tokens} serve as the representation of previous knowledge that the model has seen in a more compressed manner. 
We intend to maximize the memory size, so we assign the memory pool to every layer to significantly enlarge the memory pool.
During the generation phase, all memory tokens are used, as illustrated in Figure \ref{fig:generation}. 
The attention map is designed to
enable every token in $x$ to attend to all memory tokens. If $x$ comprises $n_x$ tokens, the attention map assumes a shape of $n_x \times (n_x + N)$, yielding a linear complexity w.r.t.~the size of the memory pool. 

\subsubsection{Self-Update Process}
\label{ssub:self_update_process}
Figure \ref{fig:injection} illustrates the self-update process. 
The goal of self-update is to ensure that \ours can always digest the latest knowledge and memorize the previously learned knowledge at its best. We discuss the self-update process in this subsection and prove in section~\ref{sec:aof} that \ours only forgets stale knowledge at an exponential decay rate with a theoretical guarantee.
When introducing new knowledge $x_c$ (in the following, we denote the new knowledge as \textit{context} $x_c$ to distinguish it from $x$ in the last section), the model must integrate $x_c$ into $\theta$ as per Eq.(\ref{eq:one_step_injection}). 
To avoid additional modules and complexities, we use the transformer $\phi$ for the update.
Ideally, the input to $\phi_l$ should be the memory pool $\theta_l$ and the hidden states $h_l$ (where $h_1$ are the word embeddings of tokenized $x_c$). We find it essential to maintain the gradient flow from both the self-update and the generation to achieve better performance (see Section ~\ref{ssub:newest_information_incorporation}). However, it is much more costly to feed the entire pool $\theta_l$ to $\phi_l$ during self-update. To solve this problem, 
we extract the last $K$ tokens of $\theta_l$ where $K<<N$ and denote these extracted tokens as $e_\theta^l$.
$e_\theta^l$ is then concatenated with $h_l$ to form the input of $\phi_l$, where $h_l$ can attend to the preceding context $e_\theta^l$.
The attention also employs an attention map of dimension
$\max(n_{x_c}, K) \times (n_{x_c} + K)$, where $n_{x_c}$ is the number of tokens in $x_c$ (Note that in Figure \ref{fig:injection} we show the case when $n_{x_c} > K$. The case when $k > n_{x_c}$ is shown in Figure \ref{fig:injection_additional} in Appendix). 
The last $K$ hidden states of the output $h_{l+1}$ are designated as ${e_\theta^l}'$. Then we drop $K$ memory tokens from the current memory pool $\theta^l$ and squeeze $\theta^l$ to the left side of the newly formed memory pool ${\theta^l}'$ where the new memory tokens ${e_\theta^l}'$ fill the right side.
In this way, ${e_\theta^l}'$ is still of dimension $d$, with the new knowledge injected into the last $K$ dimensions.

\subsubsection{Analysis of Forgetting}
\label{sec:aof}
The design of the self-update process draws inspiration from the concept of \textit{exponential forgetting} in human cognition, as described by the Ebbinghaus Forgetting Curve, and analogous to the intuition in MemoryBank~\citep{MemoryBank}. 
Through this structure, we aim to simulate exponential forgetting. In each update we drop $K$ tokens from the memory pool; statistically, we drop $K/N$ of the knowledge from the existing memory pool, which means that knowledge within the memory pool would be exponentially forgotten at a rate of $K/N$. 
Here, $N$ denotes the total memory size, while $K$ denotes the number of tokens that are used to incorporate knowledge in $x_c$. Thus, $N$ denotes the total capacity of the memory while $K$ represents the compression ratio. The smaller $K$, the more compressed the knowledge. 

After self-update, the latest knowledge is preserved entirely without any random dropping. After $N/K$ update steps, the retention ratio for knowledge injected $N/K$ steps earlier can be calculated as:
\begin{equation}\label{eq:NK}
    (1 - \frac{K}{N})^{N/K}.
\end{equation}
In Eq.~\ref{eq:NK}, with the memory pool $\theta_l$ getting larger ($N$ becomes greater) and the knowledge in $x_c$ getting more compressed ($K$ becomes smaller), we can approach the following limit: 
\begin{equation}\label{eq:NK_limit}
    \lim_{\frac{N}{K}\rightarrow \infty} (1 - \frac{K}{N})^{N/K} = 1/e,
\end{equation}
where $e$ is the natural constant, therefore, to achieve minimal forgetting, the strategy involves reducing the compression ratio (by minimizing $K$, as we essentially compress knowledge from $h_l$ into $e_\theta^{l'}$) and increasing the memory size (by maximizing $N$).

\begin{figure}[t]
    \centering
    \includegraphics[width=1.0\linewidth]{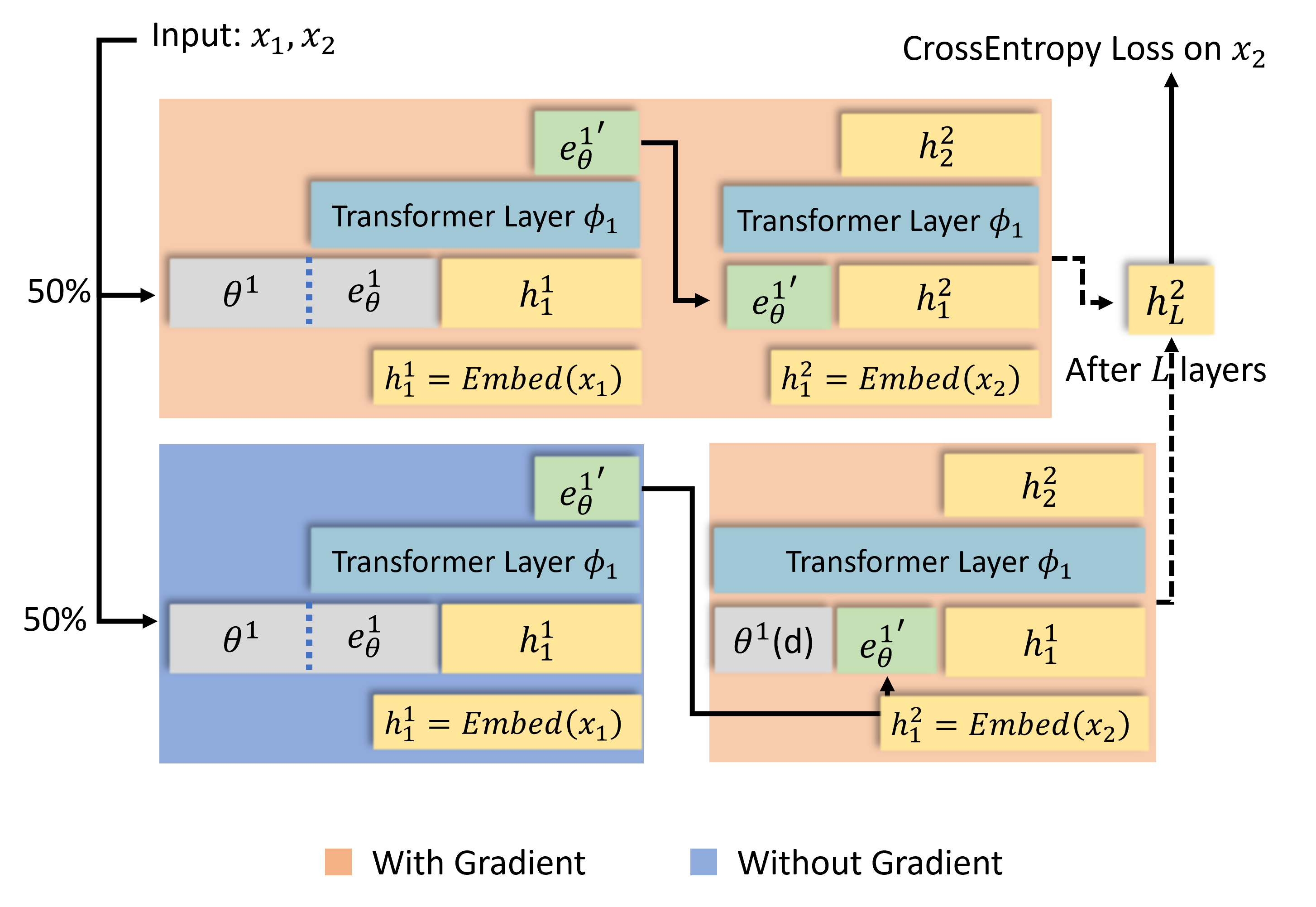}
    \vspace{-15pt}
    \caption{\textbf{Training Process for new knowledge incorporation.} 
    During training, we randomly choose one of two shown processes to proceed with 50\% probability each. The description pertains to the first layer, and the subsequent layers share an analogous procedure. 
    After sampling $(x_1,x_2)$ from the dataset, we first perform self-update with $x_1$ as depicted in the left side of both processes. Subsequently, the modified memory ${e_\theta^1}'$ is employed to predict $x_2$. 
    Of the two processes, the upper one maintains gradient flow throughout the entire process, optimizing the knowledge compression from $x_1$ to ${e_\theta^l}'$ ($l \in \{1,\cdots,L\}$). In contrast, the lower process executes the self-update without gradient. Both processes are designed to encourage the use of the knowledge in the memory pool for the prediction.
    }
    \vspace{-20pt}
    \label{fig:training-process-for-new-information-incorporation}
\end{figure}

\begin{figure}[t]
    \centering
    \includegraphics[width=.9\linewidth]{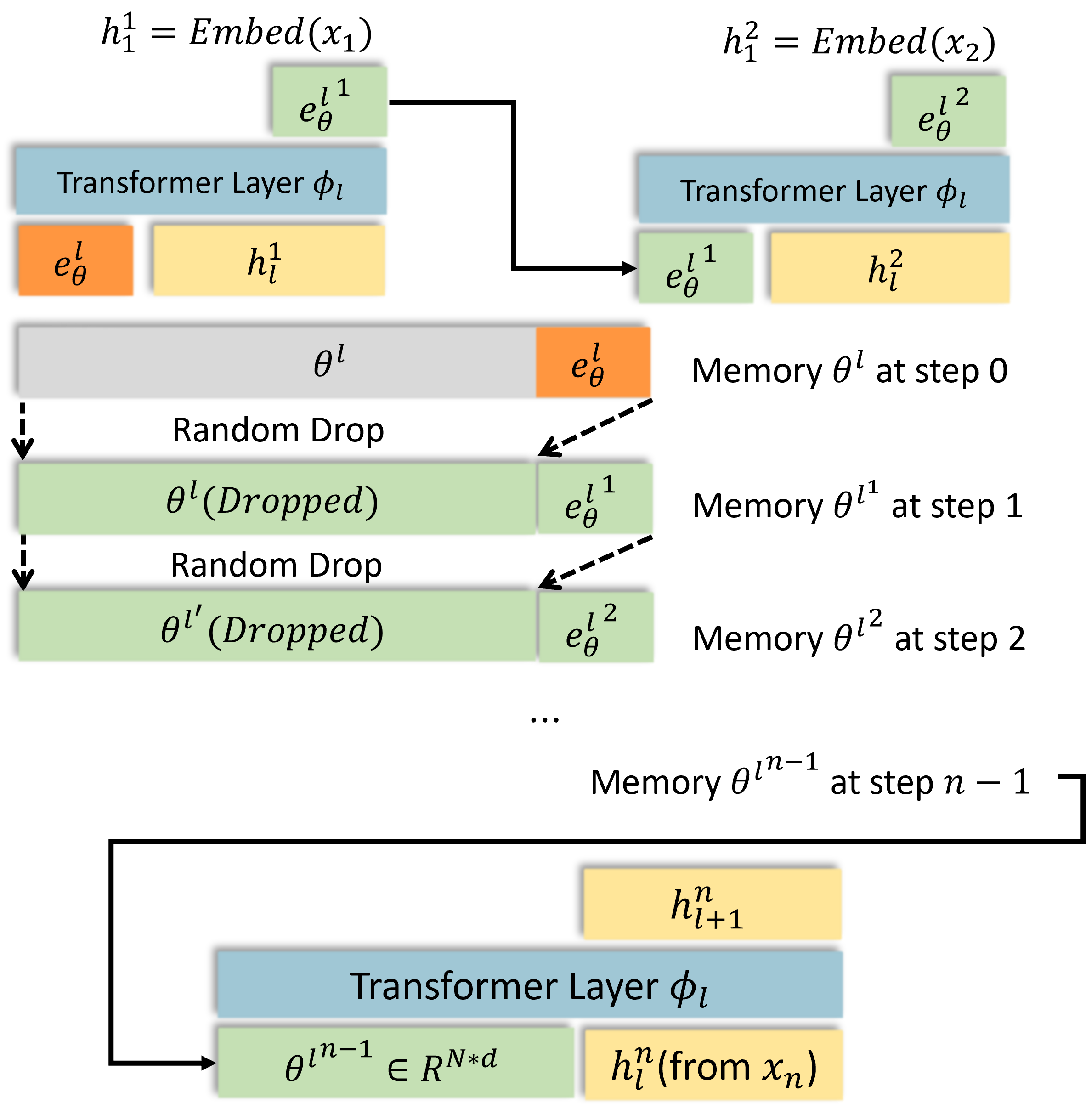}
    \caption{\textbf{Training process for continuous contexts understanding.} We only draw two self-update steps here with $x_1, x_2$ though there should be $n-1$ self-updates in this training iteration.
    We show the procedure of $l$-th layer here. At the bottom of the figure, $h_1^n$ refers to the word embeddings of $x_n$, and $h_L^n$ is used for loss value calculation. 
    Essentially we are compressing the knowledge from $x_1,\cdots,x_{n-1}$ into $\theta^{l^{n-1}}$ to predict $x_n$.}
    \vspace{-15pt}
    \label{fig:training2}
\end{figure}

\subsection{Training Strategy}
We adopt the next word prediction task to pretrain our model.
Our training methodology for \ours is strategically designed to optimize towards three core objectives discussed as follows:
\subsubsection{New knowledge incorporation}
\label{ssub:newest_information_incorporation}
The training process begins by selecting a document $d$ from the dataset, which is then divided into two segments $(x_1, x_2)$. 
Then we update the memory pool $\theta$ with $x_1$, followed by using the updated memory pool to predict $x_2$. 
The whole process is described in Figure \ref{fig:training-process-for-new-information-incorporation}. Ideally, we would design the whole process shown in the lower part of the figure with gradient enabled (see figure \ref{fig:ideal_new-information-incorporation}). 
However, this approach incurs prohibitive memory demands, especially when the memory pool is large. To mitigate this issue,
in $l$-th layer, we propose to only use ${e_\theta^l}'$ for the prediction of $x_2$ rather than the whole updated memory $\theta_l'$ when keeping the gradient flow, and use $\theta_l'$ when the self-update process with $x_1$ is performed without gradient.
In each iteration, the two aforementioned processes are randomly selected, to ensure that our model can absorb the knowledge in $x_1$ into $\theta$ and use the memory pool $\theta$ during the generation. 
\subsubsection{Enhancing continuous contexts understanding} 
\label{ssub:enhancing_continuous_contexts_understanding}
In Section \ref{ssub:newest_information_incorporation}, 
We encourage the model to understand the latest knowledge injected, where the model can make predictions based on the new memory pool $\theta'$. 
However, the model only needs the last $K$ tokens of each layer $\theta_l'$ since only ${e_\theta^{l}}'$ (the last $K$ tokens of $\theta_l'$) contains the knowledge of the last injected context. 
Thus our model may suffer from predicting the next token based on multiple injected contexts, which is essentially the long context problem. 
We propose a training routine illustrated in Figure \ref{fig:training2} to address this problem. In Figure \ref{fig:training2}, a long document is sampled and segmented into $n$ parts $(x_1, \cdots, x_n)$, with each segment being shorter than a predefined maximum length. 
The first $n-1$ segments are then sequentially injected into the memory pool $\theta$ using Eq.(\ref{eq:multi_step_injection}), resulting in $\theta_{n-1}$.
Note that this whole injection process of $(x_1, \cdots, x_{n-1})$ is executed with gradient disabled. 
Upon obtaining $\theta_{n-1}$, we calculate the cross-entropy loss on segment $x_n$. With this training procedure, we wish to enhance the model's ability to understand and process continuous contexts.

\subsubsection{Mitigating forgetting problems} 
\label{ssub:mitigating_forgetting_problems}
To address the forgetting issue, we design a task that involves contexts across multiple documents. Specifically, we sample one main document $d$ and multiple side documents $d'$ (we take one side document as an example) and split them into segments $(x_1, \cdots, x_n)$ and $(x_1',\cdots, x_n')$. The first $n-1$ segments of the main document $(x_1, \cdots, x_{n-1})$ and the side document $(x_1',\cdots,x_{n}')$ are then injected into the model sequentially. To force the model to recall the related context injected a long time ago, we make the model predict the last segment of the main document ${x_n}$. Similarly, the gradient is disabled during all the injections. We encourage the model to use the knowledge from long ago to make the prediction, thereby mitigating the forgetting problem effectively. The implementation details of this part are described in Appendix \ref{ssub:implementation_details_of_non_forgetting}.

To maintain the integrity of our model, i.e., to avoid the issue that the model may start malfunctioning after updating $\theta$ too many times, we update $\theta$ with the context after back-propagation. Specifically, we update $\theta$ with $x_1$ in Section \ref{ssub:newest_information_incorporation} and with $\{x_1,\cdots,x_{n-1}\}$ in Section \ref{ssub:enhancing_continuous_contexts_understanding} at the end of each training iteration. Intuitively, we are regularizing the distribution of ${e_\theta^l}'$ to be the same as that of $\theta_l$ to maintain integrity after arbitrarily many updates.

\vspace{-5pt}
\subsection{Model Instantiation}
We use Llama2-7b as $\phi$, consisting of $32$ layers, with a hidden dimension of $4,096$. The model we propose has $7,680$ memory tokens in every layer, meaning that $\theta \in \mathbb{R}^{32 \times 7680 \times 4096}$, comprising 1.066B parameters. 

\vspace{-5pt}
\subsection{Discussions}
\textbf{Extension to Other Architectures}: Our experimental framework involves the use of Llama2-7b as the instantiation for the function $\phi$. This selection was driven by the popularity and performance of Llama2-7b as a large language model during the development phase of our project. It is important to note, however, that the framework of our model is broadly applicable across various large language models (LLMs) that have transformer architectures with full attention mechanisms. 

\textbf{Scalability of the Memory Size}: In our main experiments, we expand the memory size to approximately 1 billion parameters. We wish to emphasize that the efficiency of the self-update process (discussed in Section \ref{ssub:self_update_process}) remains unaffected by increases in the memory pool size. This efficiency is due to the model's design, which only adopts the most recent $K$ tokens from the memory pool as the input during self-updates. Consequently, the primary scalability constraint arises from the attention mechanism between the memory tokens and the input tokens during generation (as depicted in Figure \ref{fig:generation}). As the memory pool enlarges, the computational complexity of these attention mechanisms increases \textbf{linearly} with respect to the number of tokens $N$ in the memory pool, which is because the complexity of the attention is $N\times K$. With distributed training, our framework has the potential to be scaled to significantly larger memory sizes.

\textbf{The design of Random Dropping}: Random dropping is a fairly straightforward way to keep the size of the memory pool fixed while maintaining an exponential forgetting mechanism. 
Other possible strategies include applying an exponential decay factor to the memory pool from the previous step and aggregating the decayed memory pool with the new memory. We have experimented with aggregating existing memory and new memory instead of using random dropping. However, we found that maintaining the integrity of hidden states for tokens seems to be beneficial. Aggregating hidden states often disrupts both the original and new knowledge, resulting in situations where even the knowledge injected into the memory during the last self-update process cannot be fully extracted.
In contrast, while random dropping carries the risk of forgetting previous information, it allows for the full recovery of information from the context injected during the last self-update step, as there is no random dropping applied to the new memory tokens at the last update. Therefore, we choose random dropping as we believe it provides a more natural way to integrate existing hidden states with new hidden states.

%% file: 5_experiments.tex
\vspace{-5pt}
\section{Experiments}
\vspace{-5pt}
\subsection{Evaluation Protocols}
\vspace{-5pt}
As illustrated in Section \ref{sec:introduction}, we need to evaluate \ours in the following three aspects: (1) Integration of New Knowledge: this evaluation is conducted with the model editing tasks (Section~\ref{sub:model_editing}) and QA tasks (long context QA benchmarks, Section~\ref{sub:long_context_evaluation}); 
(2) 
\wy{Knowledge Retention Ability:}
the model is evaluated with long context QA benchmarks (Section~\ref{sub:long_context_evaluation}) and our knowledge retention experiments (Section~\ref{sub:customized_experiments}); (3) Robustness: we make nearly a million updates to our memory pool and then test the functionality of our model (Section~\ref{sub:model_integrity_analysis}).

\vspace{-5pt}
\subsection{Implementation Details}
\vspace{-5pt}
We train our model on the processed version of the C4 dataset~\citep{c4} from Red-Pajama~\citep{together2023redpajama}. For the training processes in Section \ref{ssub:newest_information_incorporation}, we sample documents from the entire dataset, while the training process in Section \ref{ssub:enhancing_continuous_contexts_understanding} is based on a subset of C4 (we call this the long context subset) where all documents are of length greater than 2048. For the last part, Section \ref{ssub:mitigating_forgetting_problems}, the documents are sampled randomly from the original C4 dataset and the long context subset. The training is performed on 8 A100-80GB GPUs for three days.  

\begin{table*}[ht]
    \centering
    \caption{\textbf{Quantitative Editing Results on Llama2-7B} for ZsRE and CounterFactual Datasets. ``w/EF" means ``with editing facts", indicating the model after updating the memory pool with the new fact.}
    \label{tab:model_editing_results_llama-7b}
    \resizebox{\linewidth}{!}{
    \begin{tabular}{c|cccc|cccc}
\toprule
& \multicolumn{4}{c|}{\textbf{ZsRE Dataset}} & \multicolumn{4}{c}{\textbf{CounterFactual Dataset}} \\
\cmidrule(lr){2-5} \cmidrule(lr){6-9}
\textbf{Editor} & \textbf{Score} & \textbf{Efficacy} & \textbf{Generalization} & \textbf{Specificity} & \textbf{Score} & \textbf{Efficacy} & \textbf{Generalization} & \textbf{Specificity} \\
\midrule
Llama2-7B & 55.6 & 55.9 & 54.7 & 56.3 & 20.7 & 13.7 & 16.6 & 83.4\\
MemoryLLM-7B & 51.2 & 50.0 & 49.1 & 54.8 & 22.6 & 15.7 & 17.6 & 82.1 \\
\midrule
FT & 50.3 & 78.6 & 80.6 & 29.0 & 10.0 & 99.7 & 96.9 & 3.6\\
FT-L & 69.8 & 81.4 & 76.8 & 56.6 & 33.8 & 47.2 & 18.0 & 83.3 \\
ROME & 69.3 & 88.7 & 70.2 & 56.3 & 69.2 & 82.6 & 75.2 & 55.8 \\
IKE & - & - & - & - & 70.7 & 99.8 & 96.2 & 45.4 \\
MemoryLLM-7B (w/ EF) & \textbf{79.2} & 99.8 & 96.7 & 57.1 & \textbf{75.3} & 98.5 & 82.2 & 57.0 \\
\bottomrule
\end{tabular}}
\vspace{-10pt}
\end{table*}

\vspace{-5pt}
\subsection{Model Editing}
\label{sub:model_editing}

\vspace{-5pt}
\subsubsection{Experimental Setup}
\vspace{-5pt}

We follow the experimental setup in \cite{ROME}. The benchmarks are: 

\textbf{zsRE}~\citep{zsre}: Zero-Shot Relation Extraction(zsRE) is first used in \citet{mend, KE} for model editing evaluations.  we use the first $10,000$ records in the dataset as the evaluation set, with each record containing one factual statement. 

\textbf{CounterFactual}~\citep{ROME}: A set of more difficult false facts in which the LLMs would have low scores when prompted with these facts. Then after editing the model, the model is queried again with these facts. Each example includes the question, the original fact, and the false fact, where we aim to inject the false fact into the model. The first $2,000$ examples in this dataset are used for evaluation (following the evaluation of GPT-J in \cite{ROME}). 

Evaluation metrics include \textbf{Efficiency} (the post-edit accuracy), \textbf{Generalization} (post-edit accuracy of the paraphrased version of the factual statement), and \textbf{Specificity} (the post-edit accuracy on unrelated facts). The harmonic mean of the three metrics is reported in column \textbf{ Score}. 

We compare our model with the following baselines\footnote{We tried MEND~\citep{mend}, but the existing code is for GPT-style models. Our re-implementation of MEND on Llama2 based on the published code constantly encounters \texttt{nan}. In addition, MEND is inferior in the experiments in \cite{ROME}, thus we omit MEND in our experiments. }:
\textbf{FT}, \textbf{FT-L}~\citep{ModifyingMemories}, \textbf{IKE}~\citep{ike}, \textbf{ROME}~\citep{ROME}. The details of the baselines are described in Appendix \ref{sub:baselines_for_model_editing}.

\vspace{-5pt}
\subsubsection{Overall Performance Comparison}
The experimental results on dataset ZsRE and CounterFactual are shown in
Table \ref{tab:model_editing_results_llama-7b}. From the table, 
We observe (1) 
Our model outperforms all baseline models in both datasets, achieving the highest overall performance metrics. 
(2) While Fine-Tuning (FT) 
performs better in terms of Efficacy and Generalization, which means the model absorbs the knowledge, but it tends to lag in ``Specificity", meaning the model's knowledge of other facts is affected by the tuning.
(3) With enforced constraints (FT-L), 
the model performs better in terms of Specificity while compromising Efficacy and Generalization, representing that the knowledge is not absorbed by the model effectively.
(4) ROME, striking a reasonable balance between efficacy and specificity, compared with \ours, may fall short in the overall performance measurement. 
(5) IKE, conceptually similar to our approach by incorporating the information into the context, faces limitations in specificity, which could be ascribed to the complexity of the prompts used in the implementation, potentially disrupting the accuracy.

\vspace{-5pt}
\subsection{Long Context Evaluation}
\label{sub:long_context_evaluation}
\vspace{-2pt}
\subsubsection{Experimental Setup}
In this section, we evaluate the long-context modeling capabilities of our model. Our assessment utilizes the LongBench benchmark \citep{longbench}, specifically designed to test performance in long-context scenarios. 
Since our model has not undergone instruction fine-tuning, the baselines for comparison are also not instruction finetuned. The baselines include \textbf{Llama2-7B}: our backbone; \textbf{Longllama-3B-v1-1}\citep{fot}: The model employs contrastive learning to extend the effective context length of existing models. We adopt the 3B model here as only the 3B model is open-sourced, derived from Openllama-V2; \textbf{Openllama-V2}~\citep{openlm2023openllama}: An open-sourced reproduction of Llama. \textbf{Llama2-LongLora-7B-16k}~\citep{longlora}: A novel attention mechanism, Shift Short Attention, is proposed and used for longer context pertaining. This model, based on Llama2-7B, is extended to accommodate a 16k context length; 
\textbf{Llama2-LongLora-7B-100k}~\citep{longlora}. The same method but context length is extended to 100k. 
For 7B models, we omit the results of maximum length being $16,384$ as we encounter the out-of-memory (OOM) error even when using eight A100-80GB GPUs. This shows another advantage of \ours, as our model needs one 48 GB GPU or two 40GB GPUs to run inference regardless of the input length.

\vspace{-5pt}
\subsubsection{Overall Performance Comparison}
\begin{figure}
    \centering
    \includegraphics[width=\linewidth]{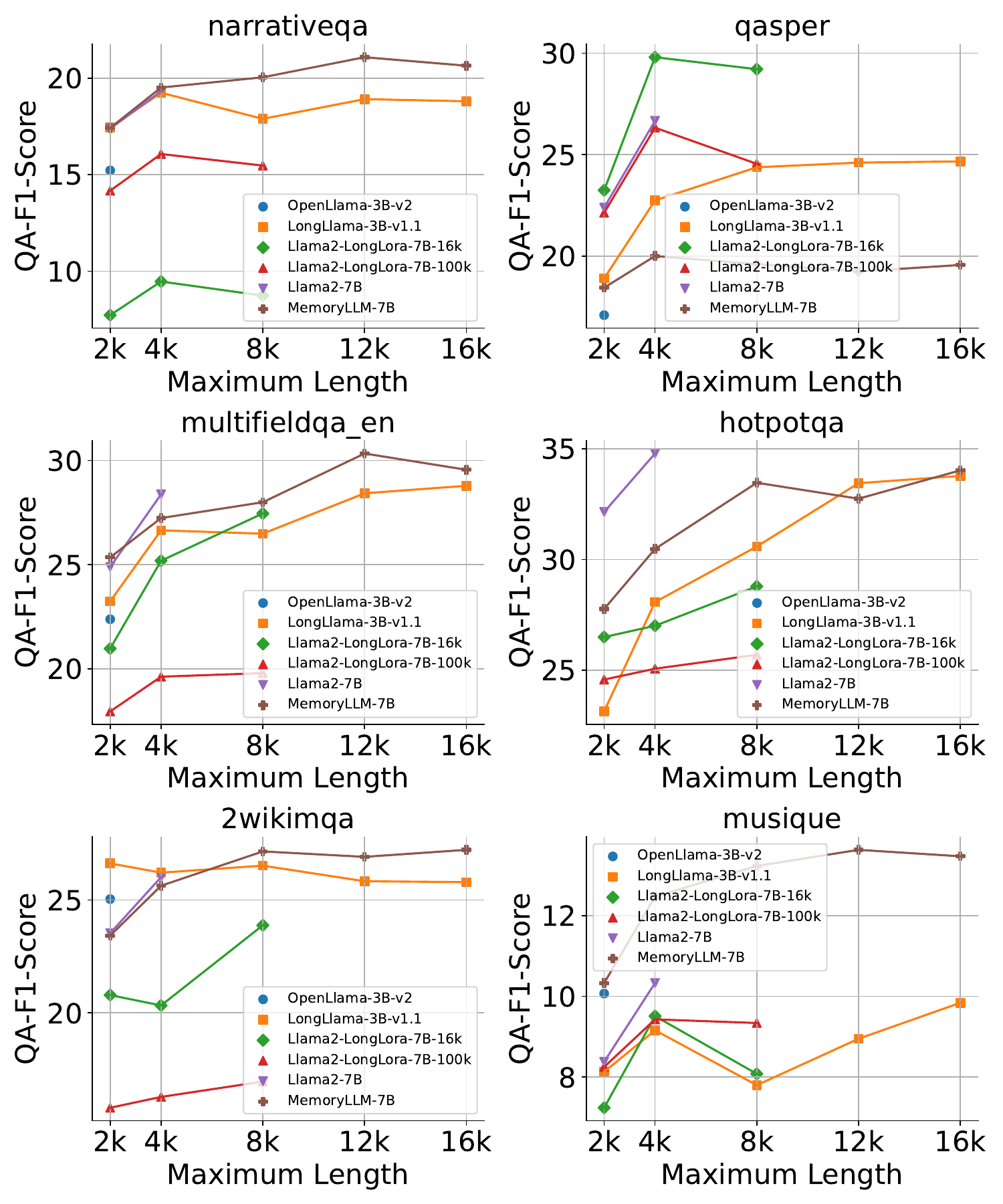}
    \vspace{-15pt}
    \caption{\textbf{Experimental Results on LongBench}. 
    The x-axis is the maximum context length for the QA task. For instance, with a maximum length of $4096$, we truncate $4096$ tokens from the given context as input to the model. The y-axis is the F1 score. 
    }
    \vspace{-10pt}
\label{fig:experimental_results_on_longbench}
\end{figure}

The results in Figure \ref{fig:experimental_results_on_longbench} reveal that: 
(1) \ours outperforms baselines in four out of six datasets when provided with extended contexts. However, a notable exception is observed in the Qasper dataset, where \ours exhibits suboptimal performance. This could be attributed to the model's training predominantly on the C4 dataset, without incorporating the arxiv dataset. \wy{Thus, the training may affect the model’s ability on scientific datasets (such as Qasper).} 
(2) As the context length grows, the performance of \ours continues to improve, demonstrating the knowledge retention ability of \ours, where the knowledge from multiple updates earlier could boost performance. 
The performance of \ours, when the context length is less than 4k, is not the same as that of Llama2-7B, which can be attributed to the subset we used for training \ours, as we do not need to use the entire dataset for pertaining Llama2-7B for our model and a subset would inevitably have distribution shift from the original dataset.

\wy{
\subsubsection{Comparison with RAG methods}
In this section, we aim to explore the role of RAG methods in QA tasks which we argue is orthogonal to \ours. 
The primary goal of MemoryLLM is to achieve self-updatable LLM where the memory module serves as the parameters that could keep updating along the inference process, whereas RAG methods aim to retrieve the most relevant piece of information from the history. Intuitively, RAG is used to conduct coarse retrieval from millions of documents, while MemoryLLM can process the retrieved documents. We use BM25 retriever to extract 4k tokens from the whole context and use MemoryLLM to process these 4k tokens to generate the answer. The results are shown in Table \ref{tab:rag}. Here MemoryLLM-7b-16k corresponds to the results in Figure \ref{fig:experimental_results_on_longbench}, and MemoryLLM-7b-all-BM25 means retrieving 4k tokens from the whole given context and using MemoryLLM to process the retrieved 4k tokens. The results show that using the BM25 retriever could enhance the model performance on certain datasets while not universally beneficial. 

\begin{table}[t]
    \centering
    \caption{The performance comparison on long context QA benchmarks of our model with and without BM25 retriever.}
    \label{tab:rag}
    \resizebox{\linewidth}{!}{
    \begin{tabular}{ccc}
    \toprule
     & \ours-16k & \ours-all-BM25 \\
    \midrule
    narrativeqa & \textbf{20.64} & 15.60 \\
    qasper & 19.57&\textbf{20.30}\\
    multifieldqa\_en & 29.56 & \textbf{33.08} \\
    hotpotqa & \textbf{34.03}& 32.27 \\
    2wikimqa & \textbf{27.22}& 24.17 \\
    musique & 13.47 &	\textbf{15.36} \\
    \bottomrule
    \end{tabular}}
    \vspace{-25pt}
\end{table}
}

\begin{figure}[t]
\centering
\subfigure[SQuAD]{\label{fig:squad_acc}\includegraphics[width=0.490\linewidth]{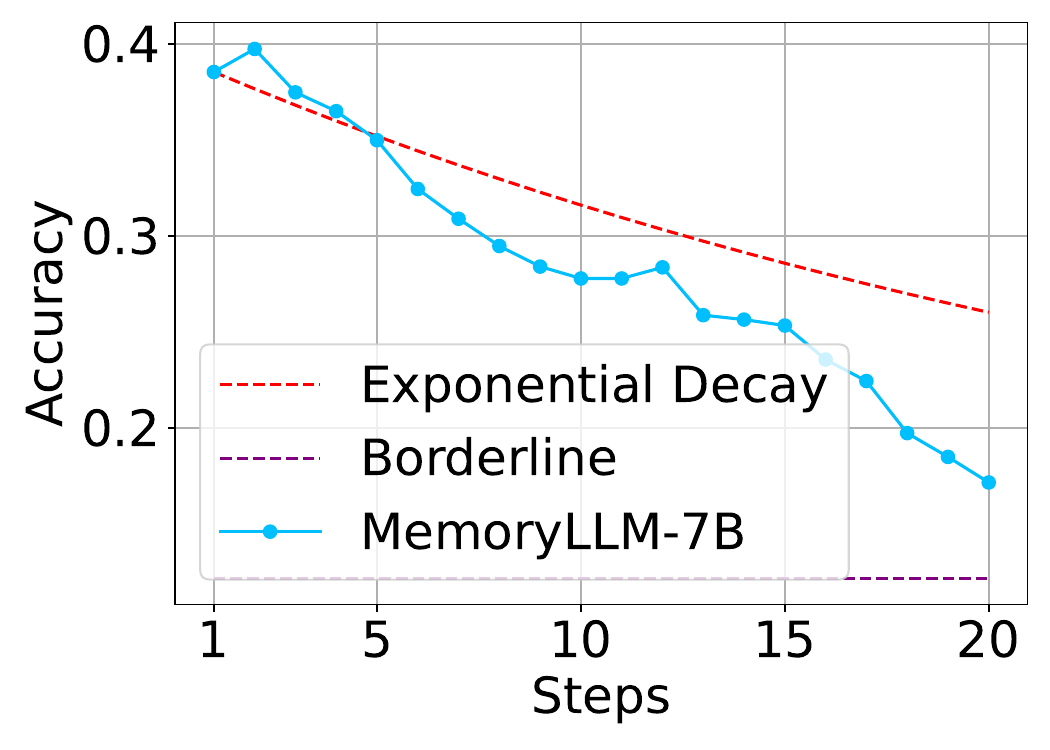}}
\hfill
\subfigure[NaturalQA]{\label{fig:naturalqa_acc}\includegraphics[width=0.490\linewidth]{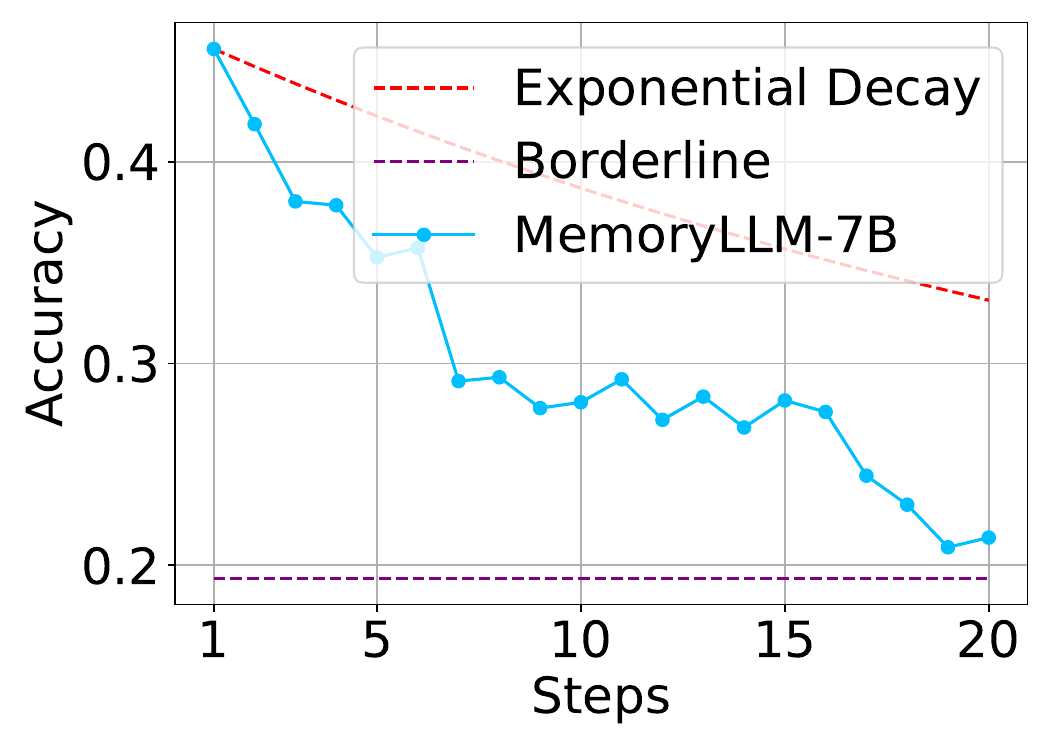}}
\caption{\textbf{Performance Comparison on SQuAD and NaturalQA.} The x-axis shows the number of updates we perform on the model, where the context that contains the knowledge to answer the question is injected in Step 1. The y-axis reveals the accuracy of the model's prediction after a certain number of updates. The accuracy is higher than the borderline indicating that the knowledge is not completely forgotten, while we wish the model to be more aligned with the exponential decay, i.e., the theoretical upper bound. }
\label{fig:performance_comparison_on_nqa_and_squad}
\end{figure}

\begin{figure}
    \centering
    \subfigure[SQuAD]{\includegraphics[width=0.49\linewidth]{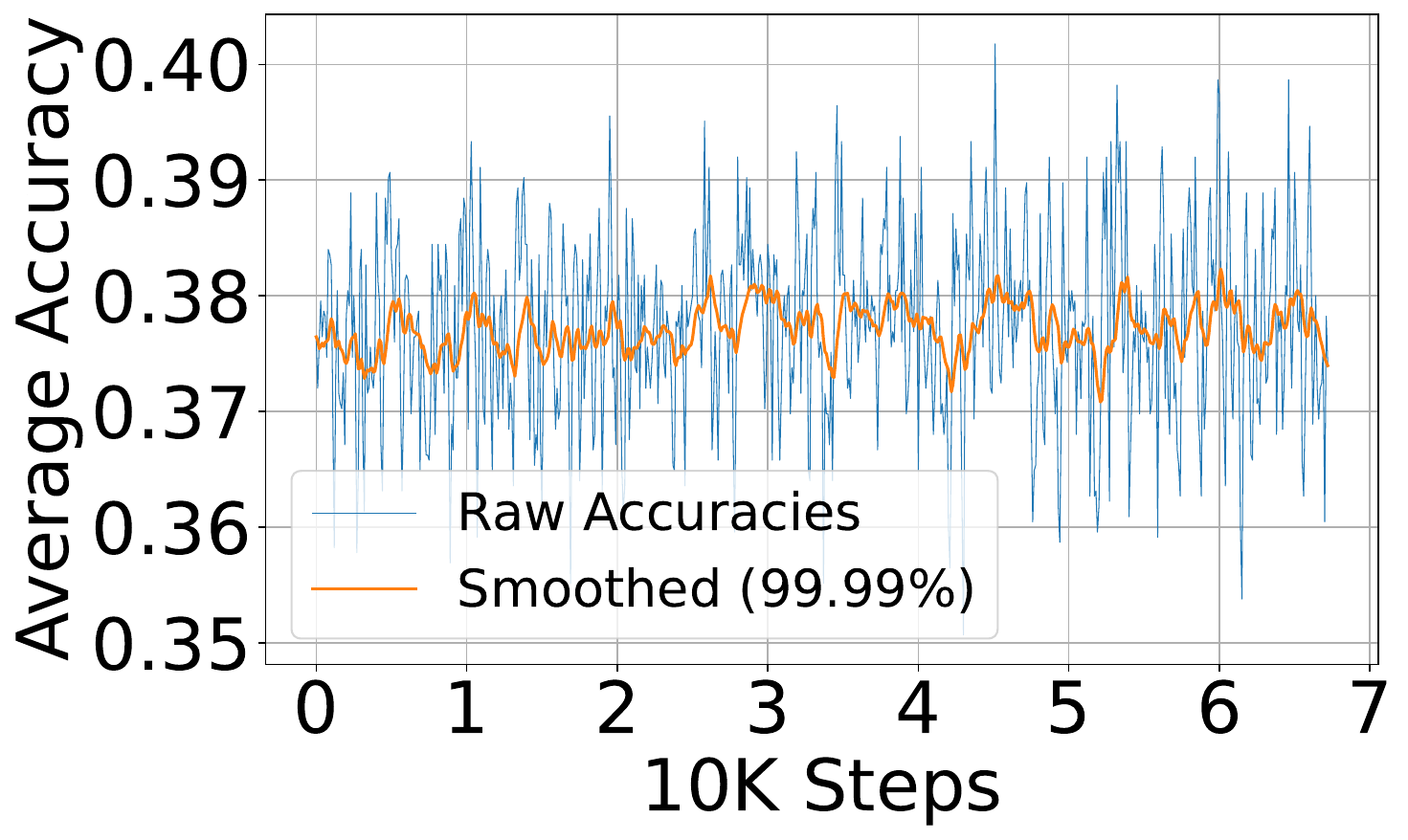}}
    \subfigure[NaturalQA]{\includegraphics[width=0.49\linewidth]{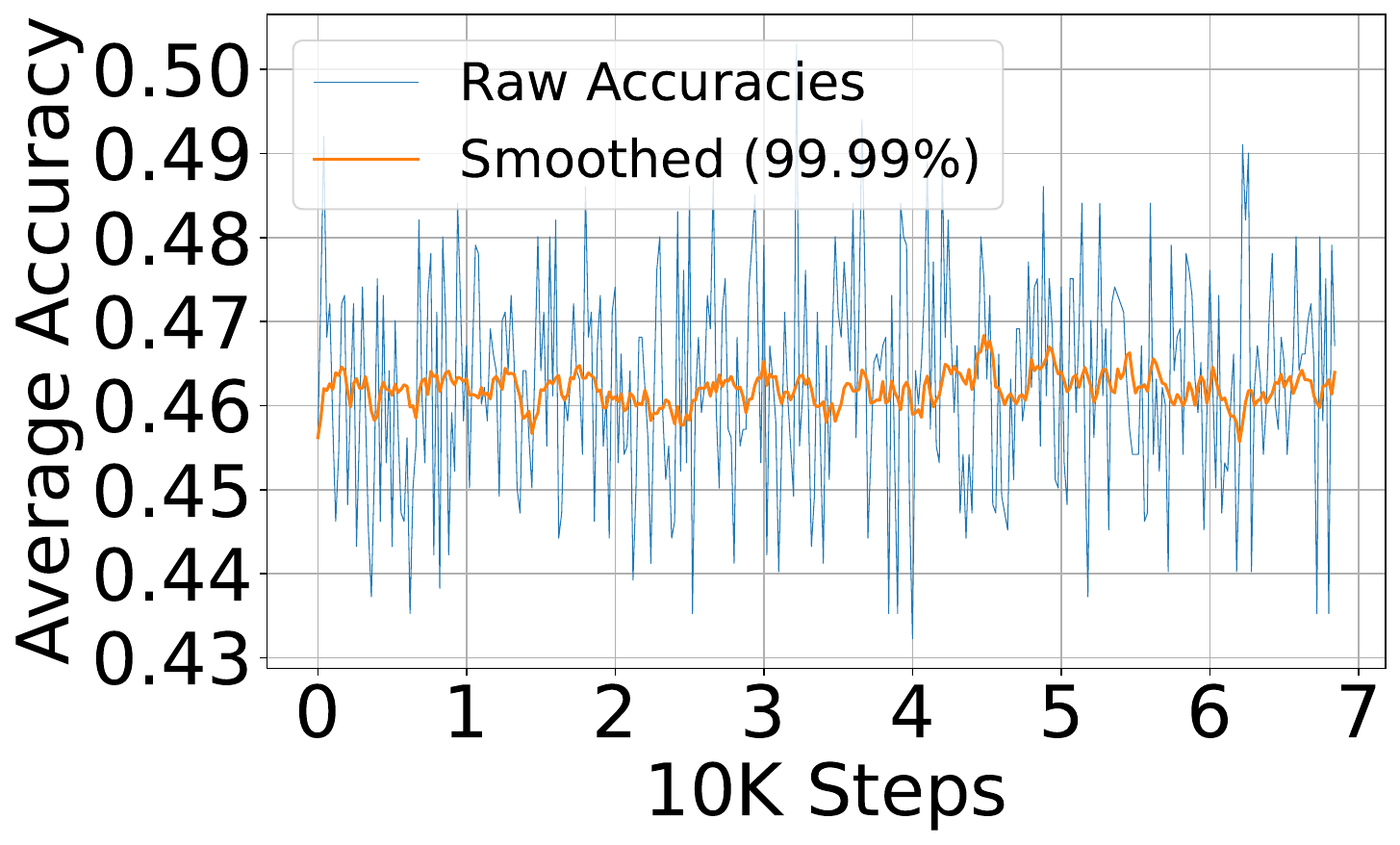}}
    \vspace{-5pt}
    \caption{\textbf{Model Integrity Check} with SQuAD and NaturalQA. We plot accuracy along the updating process as well as the exponential moving average as the \textit{Smoothed} (99.99\%) value. \textbf{We do not observe any decrease over 650k updates.}
    }
    \label{fig:model_integrity_check}
\end{figure}

\begin{figure}[ht]
\centering
\subfigure[NaturalQA Acc Percentage]{\label{fig:ablation_naturalqa_acc_percentage}\includegraphics[width=0.490\linewidth]{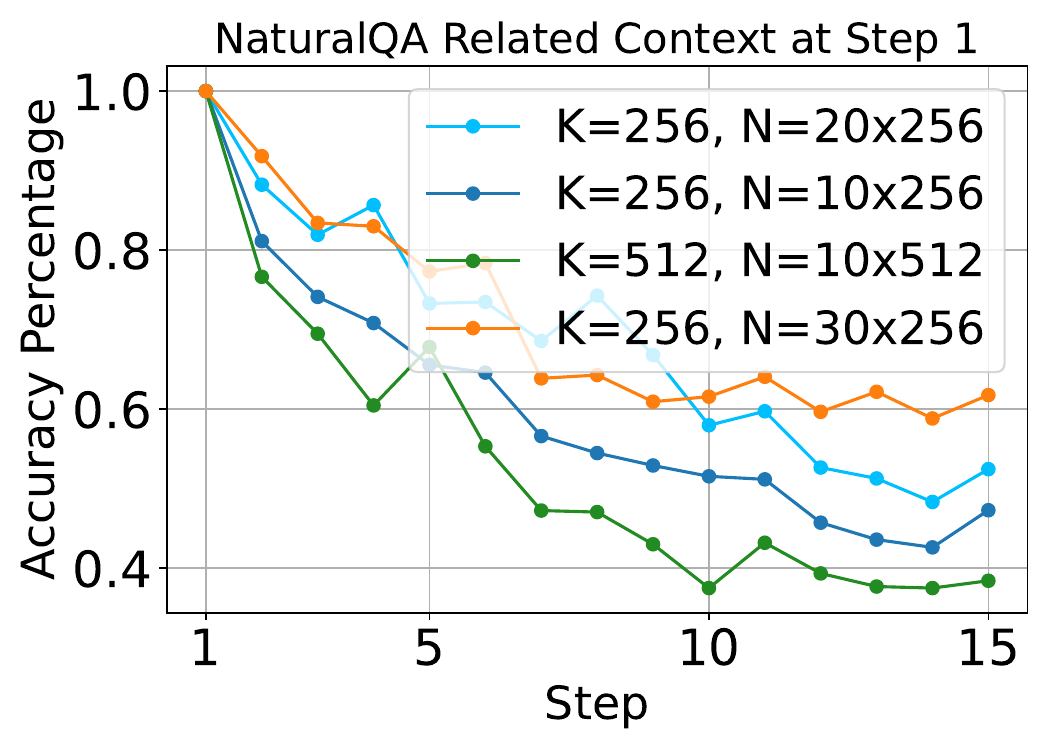}}
\hfill
\subfigure[SQuAD Acc Percentage]{\label{fig:ablation_squad_acc_percentage}\includegraphics[width=0.490\linewidth]{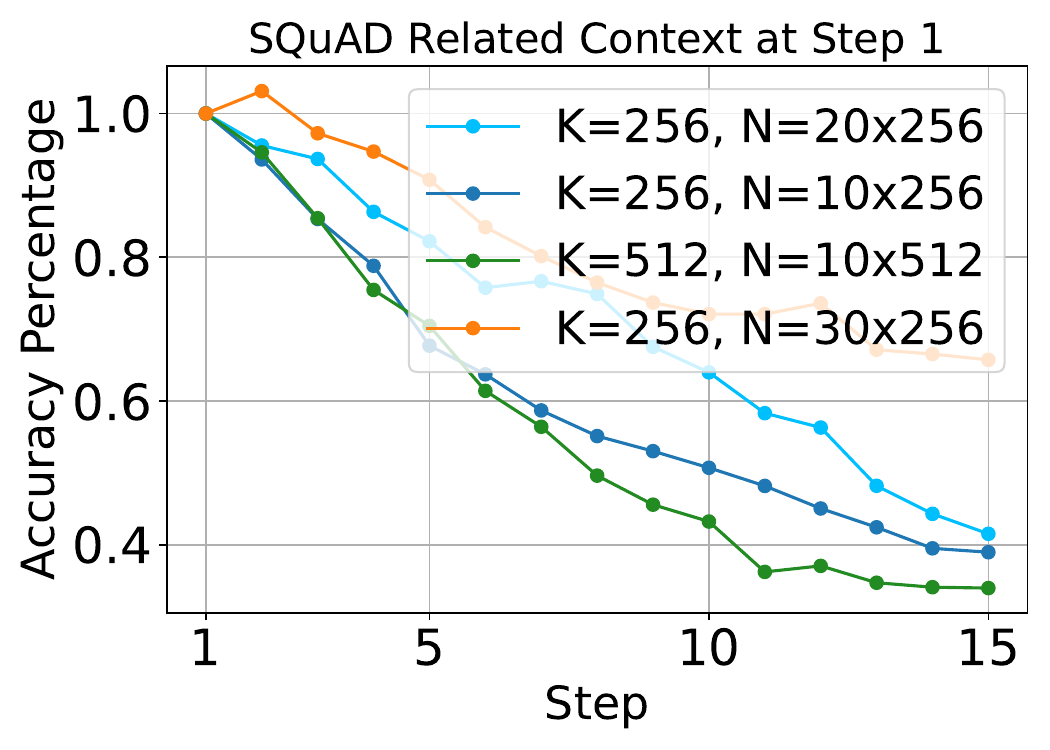}}
\subfigure[NaturalQA Accuracy]{\label{fig:ablation_naturalqa_acc}\includegraphics[width=0.490\linewidth]{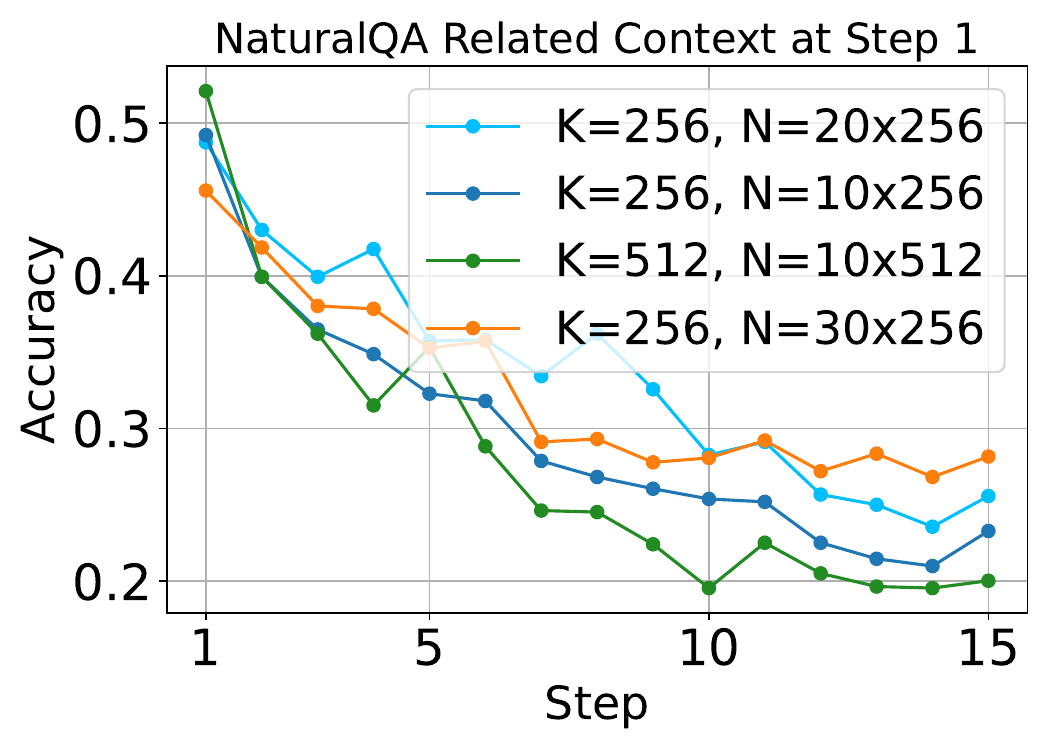}}
\hfill
\subfigure[SQuAD Accuracy]{\label{fig:ablation_squad_acc}\includegraphics[width=0.490\linewidth]{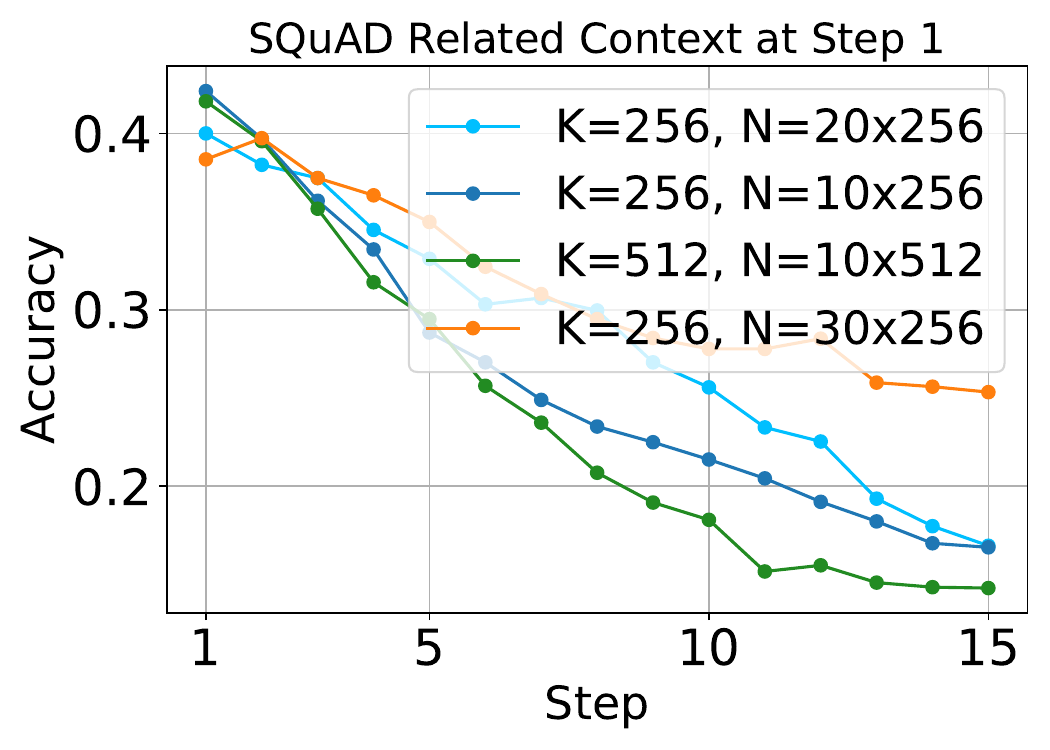}}
\caption{\textbf{Ablation Study} with our knowledge retention experiments on NaturalQA and SQuAD. All models are trained with the same setting, $30\times 256$ is our main model. 
The relevant knowledge for answering the question is injected in step 1, and the x-axis means the number of updates (steps) performed.
The top figures show the ratio of the accuracy at each step compared with the accuracy at step 1 for better visualization of knowledge retention.  }
\label{fig:ablation_study_on_nqa_and_squad}
\end{figure}

\vspace{-5pt}
\subsection{Knowledge Retention Experiments}
\label{sub:customized_experiments}

\vspace{-2pt}
\subsubsection{Experimental Setup}
The datasets are prepared as below:\\
\textbf{SQuAD}: Formatted as \texttt{(context, question, answer)}, where \texttt{context} and \texttt{question} are sentences, \texttt{answer} refers to the first answer in the list of ground-truth acceptable answers. Then we extract all the samples with \texttt{answer} shorter or equal to $3$ tokens. The model generates 10 new tokens from the prompt ``Question: \texttt{Question} Answer:". Correct predictions cover the 3-token answer within the 10 generated tokens. A total of $2,250$ samples are used for the accuracy calculation. \\
\textbf{NaturalQA}: Formatted as \texttt{(context, question, answer)}, using the long answer as the context and the short answer as the ground truth. Samples with answers of 4 tokens or less are selected. Like SQuAD, the model generates 10 new tokens, and the correct predictions cover the 4-token answer. This yields 1,004 samples for analysis. 

The results are shown in Figure \ref{fig:performance_comparison_on_nqa_and_squad}. We assess \ours's forgetting rate, comparing it against a baseline (accuracy without context injected into the memory) and a theoretical upper bound. 
Denote the accuracy at step 1 as $a_u$, and the borderline accuracy as $a_b$. Then at step $t$, we calculate the point on the curve with the following equation: 
\begin{equation}
    a_t = (a_u - a_b) * \Big(\frac{N-K}{N}\Big)^{t-1}
\end{equation}
In our instantiation, $N=7,680$ and $K=256$. Our findings indicate that the model retains knowledge even after 20 updates. However, it falls short of the exponential decay curve representing the upper bound. This gap can be attributed to the fact that even if the knowledge is partially corrupted after 20 steps of updating, it might be hard for the model to reveal the exact answer. The performance exceeding the upper bound at step 2 on the dataset SQuAD might be due to (a) the variation of inference and (b) dropping a small part of the memory may not affect the model predicting the words, while the exponential curve would drop.

\subsection{Model Integrity Analysis}
\label{sub:model_integrity_analysis}
To illustrate the integrity of our model, 
We update our model with NaturalQA and SQuAD mentioned in Section \ref{sub:customized_experiments}. 
Each time we go through the whole dataset, we shuffle the dataset and inject it into the memory again. In this way, we can simulate infinite updates. Then during the updating process, we track if our model could answer the question related to the most recent context, obtaining a long binary array that indicates whether our model successfully answers the question related to the most recently injected context. With this binary array, we calculate the average accuracy of the last $2,250$ samples for SQuAD and the last $1,004$ samples for NaturalQA. The results are shown in Figure \ref{fig:model_integrity_check}. We continue running for up to $650,000$ steps for $3$ days. As shown in the figure, there is no sign of decreasing in accuracy even after the $650,000$ steps, demonstrating the integrity of our model. From this observation, we argue that our model could be potentially updated for arbitrarily many times without affecting the functioning ability.

\vspace{-5pt}
\subsection{Ablation Study}
\wy{
\subsubsection{Ablation Study of different $K$ and $N$}
}
In this section, we study the effects of different $K$ and $N$ in Eq.(\ref{eq:NK}) with our knowledge-retention experiments. 
Our primary goal is to explore the forgetting ratio when the model has different memory sizes ($N$) and numbers of tokens to store the new knowledge ($K$). We vary $N$ to be $\{10 \times 256, 20 \times 256, 30 \times 256\}$, and $K$ to be $\{256, 512\}$. We also tried $K=128$, but to find that the accuracy is much worse than the other settings \wy{(the step 1 accuracy of NQA and SQuAD under the setting $K=128$ are 0.34 and 0.25, respectively)}, we omit this setting here. The results shown in Figure \ref{fig:ablation_study_on_nqa_and_squad} reveal that (1) When $K$ is fixed, with greater $N$, the forgetting ratio is smaller; (2) When $N$ is fixed, with smaller $K$ ($10\times512$ vs. $20\times256$, the latter yields better knowledge-retention ability), the forgetting ratio becomes smaller. 
These experiments support our intuition and show that with the improvement of $N/K$, we can enable better knowledge-retention ability. 
\wy{
\subsubsection{Ablation Study of the Model Structures}
In our main experiments, we train the model with the memory tokens augmented in every layer. To study the necessity of this design, we tried the following several settings: (1) Augment only one layer in the model with memory tokens; (2) Augment the last half of the layers in the model with the memory tokens (this design is inspired by Figure 6(a) in \citet{fang2024unimem}). Then we find that design (1) leads to almost zero improvements with the context compared to the performance without the context, which means augmenting only one layer is almost useless. For design (2), We record the accuracies after injecting the context for one step: NaturalQA: 0.39, SQuAD: 0.22. For reference, the accuracy of NaturalQA and SQuAD in Figure \ref{fig:ablation_study_on_nqa_and_squad} at step 1 is 0.46 and 0.39 respectively. This shows that having the memory tokens in both the first half and the second half layers is necessary for better performance.
}

%% file: 2_related_work.tex
\section{Related Work}
\subsection{Memory based methods}
Previous memory-based methods share certain similarities with \ours. Among these methods, some use an external encoder to inject knowledge into the memory pool, such as the Memory Network~\citep{MemoryNetwork}, which focuses on rectifying the forgetting problems in RNNs. Follow-up work \citet{E2EMN} computes the weighted sum of the entire memory pool as the representative vector of the memory. Others use the language model itself as the encoder to update the memory. Memory Transformer~\citep{memoryTransformer} and RMT~\citep{RMT} propose to add memory tokens when reading the contexts, where the memory pool is up to $20$ tokens. EMMA~\citep{Memory-Enhanced-Transformer} has a slightly larger memory pool, which is the size of the chunk when injecting the contexts into the memory. These fixed-sized memory pools show promising results, although performance is limited by the size of the memory pool. This also shows the challenges of expanding memory and incorporating information without disturbing the original capability of the model. 

Other memory-based methods integrate the memory pool with unfixed size, where different forgetting mechanisms are adopted to handle the ever-growing problem. In this case, the memory pool would be in the form of (1) hidden states, such as \citep{GlobalMemoryTransformer} and MemoryBank~\citep{MemoryBank}; (2) key-value pairs, represented by KNN-LM~\citep{kNNLM}, LONGMEM~\citep{LongMEM}. (3) vectors in hidden space. This involves the image captioning task~\citep{MeshedTforIC} and Memformer~\citep{Memformer}. (4) raw texts. RET-LLM~\citep{ret-llm} proposes to save the knowledge with triplets into the memory and then use API query to retrieve related information in the memory given the context. 
These methods have a more flexible memory pool. However, the memory pool might be redundant in terms of the stored knowledge.

\subsection{Downsteam Tasks}
As \ours has a large memory pool that can be used to store knowledge, it could be used for downstream tasks such as model editing and long context tasks. 

For model editing tasks~\citep{LLMEditing}, MEND~\citep{mend} and ROME~\citep{ROME} propose to modify the parameters of the LLM with the new given fact. During inference, MEND needs back-propagation and ROME requires the optimization for new MLP weights, while \ours, regarding the memory pool as part of the model parameters, could directly update the memory pool to store new facts. IKE~\citep{ike} proposes to simply put the new facts in context, which is straightforward and intuitively similar to \ours in terms of this task. However, IKE would encounter the same problem as long context methods, i.e., the ever-growing contexts. 

For Long context tasks, representative methods can be categorized as follows: (1) Efficient Attention such as Longformer~\citep{longformer}, Linformer~\citep{linformer}, LongNet~\citep{longnet}, (2) Positional Encoding like Alibi~\citep{alibi}, Positional Interpolation~\citep{PositionalInterpolation} and Extrapolation~\citep{lextransformer}, (3) Finetuning with longer context~\citep{Llama2Long, fot}, (4) Memory-based methods~\citep{LongMEM, RMT, Memformer}. Among all these categories, \ours could fit into the fourth category where long contexts are absorbed into the memory, which is used for future prediction.

%% file: 6_conclusion.tex
\section{Conclusion and Future Work}
In this paper, we propose \ours, a language model
consisting of a transformer and a huge memory pool within the latent space of the transformer, which serves as the self-updatable parameters of the model. 
\ours
can perform self-updates on the memory with new knowledge, enabling effective knowledge incorporation and slow forgetting of previous knowledge.
Comparisons against baselines for model editing and long context, together with a dedicated customized evaluation for knowledge retention analysis, demonstrate the superiority of \ours in effectively absorbing new knowledge and knowledge retention ability. 
In the future, it is of interest to extend the memory size as well as increase the compression rate, i.e., using fewer memory tokens during self-update to store the new knowledge. 
In addition, we aim to extend \ours to be multimodal, as the memory tokens of \ours may be suitable for storing multimodal knowledge.

%% file: 8_statement.tex
\section*{Impact Statement}
This paper presents work that aims to advance the field of Natural Language Processing, specifically the Large Language Models. There are many potential societal consequences of our work associated with LLMs, such as AI safety and reliability. Beyond LLMs, we feel no other consequences must be highlighted here.

%% file: 7_appendix.tex
\section{Details in Methodology}
\subsection{Self-Update Process}
\begin{figure}[h!]
    \centering
    \includegraphics[width=0.6\linewidth]{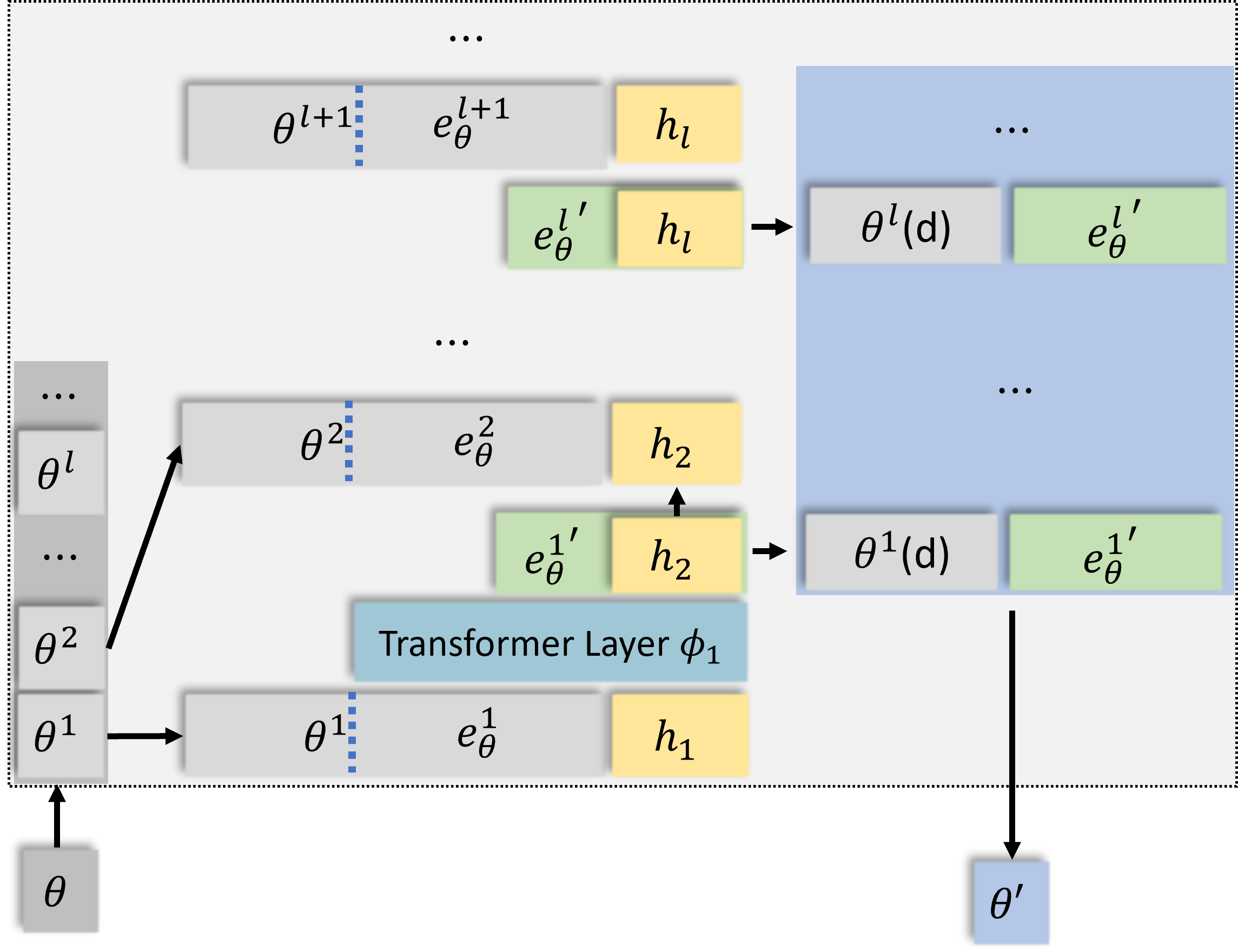}
    \caption{Self-Update process when the number of tokens is smaller than the number of memory tokens needed.}
    \label{fig:injection_additional}
\end{figure}

In Section \ref{ssub:self_update_process}, we illustrate the self-update process with the scenario of the input context $x_c$ having more than $K$ tokens. As for the case when $x_c$ has less than $K$ tokens, we draw the process in Figure \ref{fig:injection_additional}. As shown in this figure, we input $e_\theta^l$ and $h_l$ into the transformer layer $\phi_l$ to obtain ${e_\theta^l}'$ where the last $n_{x_c}$ tokens are passed into the next layer. 

\subsection{Training Strategy for New Knowledge Incorporation}
\begin{figure}
    \centering
    \includegraphics[width=0.6\linewidth]{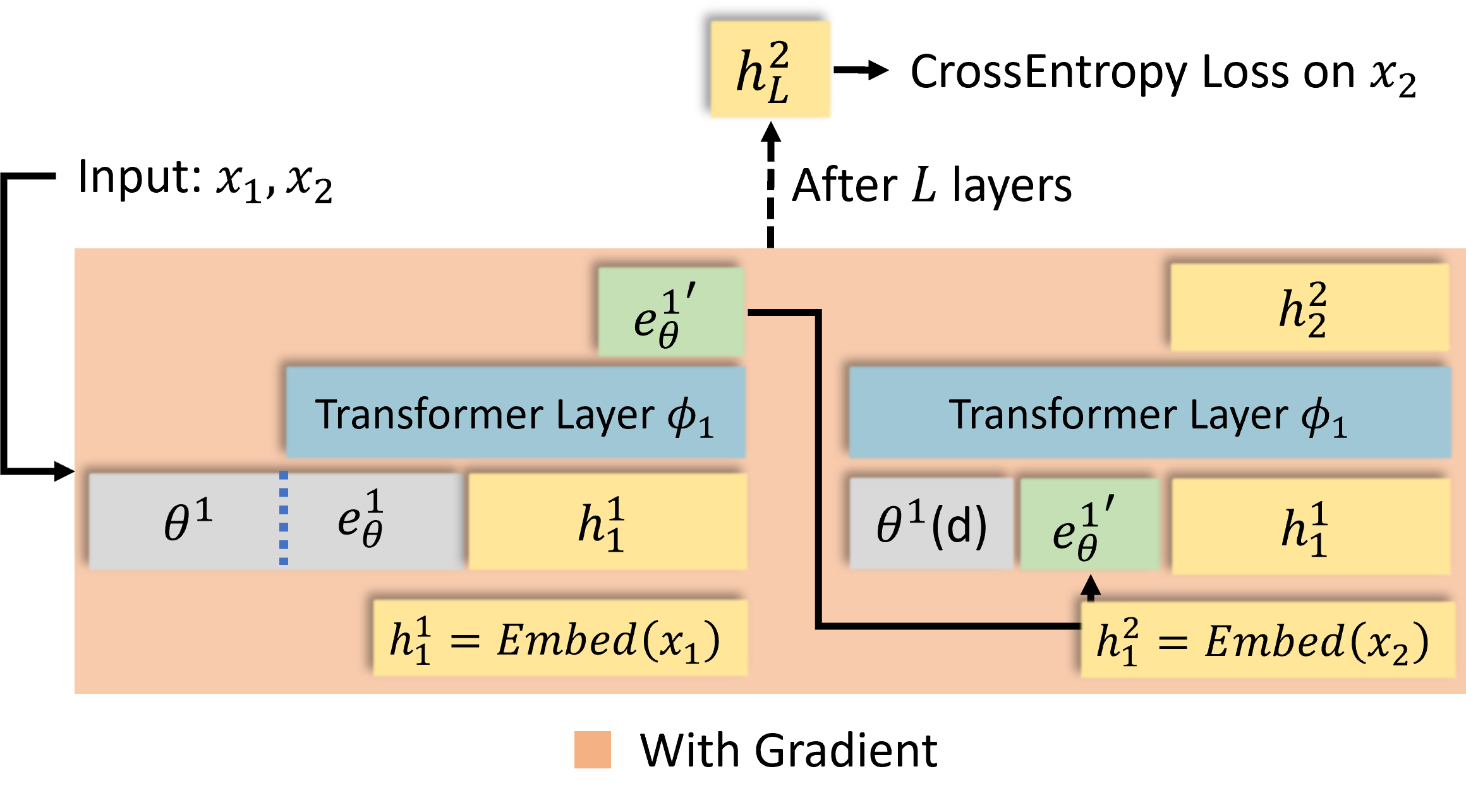}
    \caption{Ideal Training Routine for Latest Knowledge Incorporation}
    \label{fig:ideal_new-information-incorporation}
\end{figure}
As shown in Figure \ref{fig:ideal_new-information-incorporation}, compared with Section \ref{ssub:newest_information_incorporation}, the ideal case is to perform the whole process, i.e., self-update and the prediction on $x_2$, with gradient flow, so that the cross-entropy loss could be backpropagated to $x_1$. However, this would induce unaffordable memory consumption, thus we decompose this process into two processes in Figure \ref{fig:training-process-for-new-information-incorporation}. 

\section{Implementation Details}
\subsection{Details for Mitigating Forgetting Problems}
\label{ssub:implementation_details_of_non_forgetting}
As mentioned in Section \ref{ssub:mitigating_forgetting_problems}, we need to sample one main document $d$ = $\{x_1,\cdots,x_n\}$ and multiple side documents and inject all the side documents into the memory after the injection of $\{x_1,\cdots,x_{n-1}\}$, then we calculate the loss on $x_n$ to update the model. However, instead of sampling multiple documents at each step, we develop a more efficient strategy during training. We provide the pseudo-code in Algorithm \ref{alg:training_strategy}. 

\begin{algorithm}[H]
\caption{Training Strategy for Mitigating Forgetting Problems}
\label{alg:training_strategy}
\begin{algorithmic}[1]
\REQUIRE Training data $\mathcal{D}$;
\STATE Initialize the indicator $r_0=1$, $l=0$;
\STATE Initialize the cache $x_{cache}$ = None;
\FOR{$d \in \mathcal{D}$}
\STATE $n = $ the number of contexts in $d$;
\STATE $\{x_1, \cdots, x_n\} = d$;
\IF{$r_0 == 1$ or $l==0$}
    \STATE $r$ = 0; 
\ELSE
    \STATE $r$ = Random(0, 2);
\ENDIF
\IF{$r==0$ and $r_0==0$}
    \STATE Inject $\{x_1,\cdots,x_{n-1}\}$ into the memory pool; 
    \STATE Calculate the cross-entropy loss on $x_n$ and update the model;
    \STATE $l+=n$;
\ELSIF{$r==0$ and $r_0==1$}
    \STATE Inject $\{x_1,\cdots,x_{n-1}\}$ into the memory pool; 
    \STATE Calculate the cross-entropy loss on $x_n$ and update the model;
    \STATE $x_{cache} = x_n$;
    \STATE $l+=n$;
\ELSIF{$r==1$}
    \STATE Calculate the cross-entropy loss on $x_{cache}$ and update the model; 
    \STATE $l=0$;
\ENDIF
\STATE $r_0=r$;
\ENDFOR
\end{algorithmic}
\end{algorithm}
Note that at every step, we inject the knowledge into the memory pool, thus after a random number of steps, the useful knowledge for predicting $x_{cache}$ must be somewhere in the memory pool, we need to encourage the model to extract the relevant knowledge. If the model could extract the knowledge from the memory that was injected long ago, we could mitigate the forgetting problems.

\section{Additional Experiments}
\subsection{Baselines for Model Editing}
\label{sub:baselines_for_model_editing}
We introduce the details of the baselines for the model editing experiments here: 

\textbf{FT} (Finetuning): which applies Adam with early stopping at one layer to finetune the model on the given fact.

\textbf{FT-L} (Constrained Finetuning)~\citep{ModifyingMemories}: a parameter-space $L_{\infty}$ norm constraint is imposed on the weight changes. 

\textbf{IKE} (In-context knowledge editing)~\citep{ike}: The facts used to edit the model are saved in the contexts, which are inputted into the model during inference. This method is only implemented on CounterFactual so we compare our model with it on the CounterFactual benchmark. 

\textbf{ROME} (Rank-One Model Editing)~\citep{ROME}: After identifying that MLPs in LLMs are the major modules for saving knowledge, ROME proposes to alter the MLP matrix by regarding the matrix as a key-value store and then insert a new key-value pair into the matrix, obtaining a new one that contains the injected information.